\documentclass{article}
\PassOptionsToPackage{numbers, compress}{natbib}

\usepackage[preprint]{neurips_2023}
\usepackage{subfig}
\usepackage{caption}
\usepackage{graphicx}
\usepackage{booktabs}       %
\usepackage[utf8]{inputenc} %
\usepackage[T1]{fontenc}    %
\usepackage{hyperref}       %
\usepackage{url}            %
\usepackage{booktabs}       %
\usepackage{amsfonts}       %
\usepackage{nicefrac}       %
\usepackage{microtype}      %
\usepackage{xcolor}         %
\usepackage{amsmath} 

\title{Adaptive Window Pruning for \\Efficient Local Motion Deblurring}

\author{
    Haoying Li\textsuperscript{\rm 1,\rm 2},
    Jixin Zhao\textsuperscript{\rm 1},
    Shangchen Zhou\textsuperscript{\rm 1},
    Huajun Feng\textsuperscript{\rm 2},
    Chongyi Li\textsuperscript{\rm 1}\thanks{Corresponding author},
    Chen Change Loy\textsuperscript{\rm 1}\\
    \\
    \textsuperscript{\rm 1}S-Lab, Nanyang Technological University \\ 
    \textsuperscript{\rm 2}State Key Laboratory of Extreme Photonics and Instrumentation
\\
    \texttt{\small \{n2207928h,ZHAO038,s200094,chongyi.li, ccloy\}@ntu.edu.sg}$\,\,\,\,$fenghj@zju.edu.cn\\
    {\tt\small \url{https://leiali.github.io/LMD-ViT_webpage/index.html}}
}

\begin{document}

\maketitle

\begin{abstract}
Local motion blur commonly occurs in real-world photography due to the
mixing between moving objects and stationary backgrounds during exposure. 
Existing image deblurring methods predominantly focus on global deblurring, inadvertently affecting the sharpness of backgrounds in locally blurred images and wasting unnecessary computation on sharp pixels, especially for high-resolution images.
This paper aims to adaptively and efficiently restore high-resolution locally blurred images.
We propose a local motion deblurring vision Transformer (LMD-ViT) built on adaptive window pruning Transformer blocks (AdaWPT). 
To focus deblurring on local regions and reduce computation, AdaWPT prunes unnecessary windows, only allowing the active windows to be involved in the deblurring processes. The pruning operation relies on the blurriness confidence predicted by a confidence predictor that is trained end-to-end using a reconstruction loss with Gumbel-Softmax re-parameterization and a pruning loss guided by 
annotated blur masks.
Our method removes local motion blur effectively without distorting sharp regions, demonstrated by its exceptional perceptual and quantitative improvements compared to state-of-the-art methods. 
In addition, our approach substantially reduces FLOPs by 66\% and achieves more than a twofold increase in inference speed compared to Transformer-based deblurring methods.
We will make our code and annotated blur masks publicly available.

\end{abstract}

\vspace{-1mm}
\section{Introduction}
\begin{figure}[t]
    \centering
    \includegraphics[width = \textwidth]{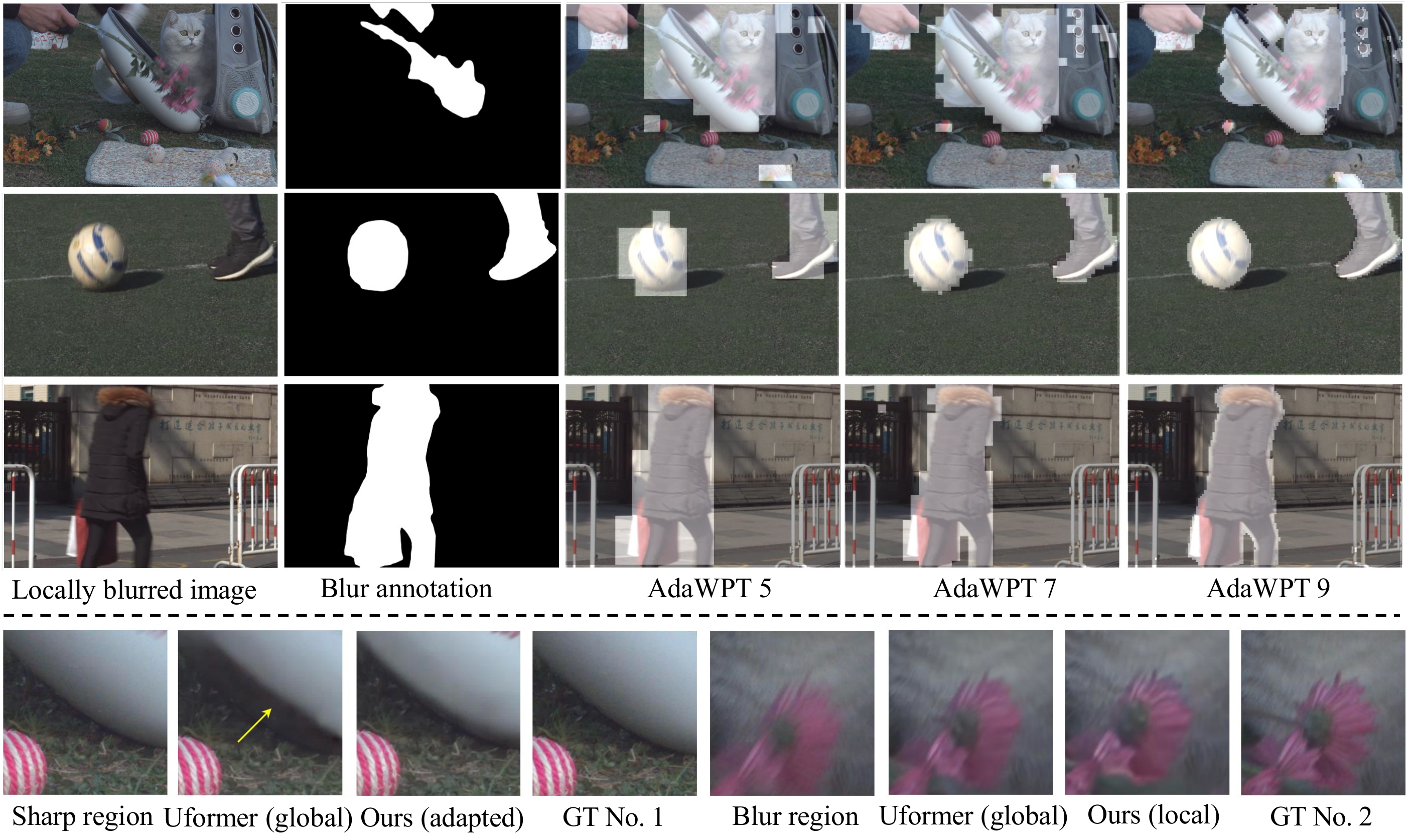}
    \caption{We introduce LMD-ViT, a Transformer-based local motion deblurring method with an adaptive window pruning mechanism. We prune unnecessary windows based on the predicted blurriness confidence supervised by our blur region annotation. In this process, the feature maps are pruned at varying levels of granularity within blocks of different resolutions. The white masks in AdaWPT 5 to 9 denote tokens to be preserved and regions without white masks are pruned. Unlike global deblurring methods 
    that modify global regions \citep{wang2022uformer,zamir2022restormer}, LMD-ViT performs dense computing only on the active windows of blurry regions. Consequently, local blurs are efficiently removed without distorting sharp regions. 
    }
    \label{fig:motivation}
    \vspace{-4mm}
\end{figure}
Contrary to global motion blur, which typically affects an entire image \citep{zhang2018learning}, local motion blur is confined to specific regions within an image. Such local blur is generally the result of object movements captured by stationary cameras \citep{li2022real,schelten2014localized}. Applying global deblurring methods to images featuring local motion blur inevitably introduces unwanted distortions in regions that were originally sharp, as illustrated in \autoref{fig:motivation}. Moreover, the processing of sharp regions, which is not required in this context, leads to unnecessary computational expenditure. This wastage becomes particularly noticeable when dealing with high-resolution inputs.

To achieve long-range pixel interactions, Transformer \citep{vaswani2017attention} has been proven to be successful and introduced in several image restoration problems \citep{liang2021swinir,zamir2022restormer, wang2022uformer}. 
It utilizes a self-attention mechanism (SA) to provide pixels with a sense of global information and exhibits better performance than CNN-based methods. 
However, Transformers usually require a high memory footprint and long inference time, particularly for processing high-resolution images. 
To alleviate computation costs, methods are proposed such as window-division strategies \citep{wang2022uformer} and token halting strategies \citep{rao2021dynamicvit,yin2022vit}. %

Existing local motion deblurring methods, such as LBAG \citep{li2022real}, address the issue by detecting local blur regions for targeted processing. While LBAG uses a gate structure to mitigate the deblurring impact on non-blurred regions, 
it still involves unnecessary computations as the entire image is processed by the network. 
In addition, the method's reliance on a Convolutional Neural Network (CNN) architecture leads to limitations, as the interactions between the image and convolution kernels are content-independent and ill-suited for modeling long-range dependencies.
The Transformer architecture \citep{vaswani2017attention}, which excels at long-range pixel interactions, has been successfully applied in several image restoration problems \citep{liang2021swinir,guo2023shadowformer,wang2022uformer,zamir2022restormer}. However, Transformers tend to require substantial memory and extend inference time, especially when processing high-resolution images. To mitigate these computational demands, strategies such as window-division~\citep{wang2022uformer,yang2021focal} and token-reducing \citep{bolya2022token,liang2022not,meng2022adavit,rao2021dynamicvit,yin2022vit} have been proposed. We explain the background of our

Drawing inspiration from preceding work, in this paper, we propose a U-shaped local motion deblurring vision Transformer (LMD-ViT) with adaptive window pruning Transformer blocks (AdaWPT) as its core component. 
Our proposed LMD-ViT is the first to apply sparse vision Transformer to the local motion deblurring task.
The core block, AdaWPT, focuses on locally blurred regions rather than global regions, which is made possible by removing windows unrelated to blurred areas, surpassing prior state-of-the-art methods in both deblurring performance and efficiency.
Specifically, we first train a confidence predictor which is able to automatically predict the confidence of blurriness of feature maps. 
It is trained end-to-end by a reconstruction loss with Gumbel-Softmax re-parameterization, and a pruning loss guided by our elaborately annotated local blur masks.
We then design a decision layer that provides binary decision maps in which ``1'' represents the kept tokens in blur-related regions that require processing during inference while ``0'' represents the abandoned tokens in the other regions that can be removed. 
We also propose a window pruning strategy with Transformer layers. In detail, we apply window-based multi-head self-attention (W-MSA) and window-based feed-forward layers rather than enforcing these Transformer layers globally. 
Only the selected windows are forwarded to these window-based Transformer layers, preventing unnecessary distortion of sharp regions while also reducing computations. 
To further enhance content interactions, AdaWTP employs shifted window mechanism \citep{liang2021swinir,liu2021swin} and position embeddings \citep{wang2022uformer} among the Transformer layers. 
Furthermore, we insert AdaWTP in LMD-ViT under different scales and receptive fields.
Therefore, AdaWTP conducts coarse pruning of windows in low-resolution layers and more refined pruning in high-resolution layers, as shown in \autoref{fig:motivation}, which achieves a balance between computational complexity and deblurring performance. 

To summarize, our main contributions are 1) the first sparse vision Transformer framework for local motion deblurring, LMD-ViT, focusing computation on localized regions affected by blur and achieving efficient and effective blur reduction without causing unnecessary distortion to sharp regions; 2) an adaptive window pruning Transformer block (AdaWPT), which prunes unnecessary windows according to a decision layer as well as a confidence predictor, and speeds up Transformer layers by a novel window pruning strategy; 3) carefully annotated local blur masks for ReLoBlur dataset \citep{li2022real}, which improve the performance of local deblurring methods. 
\vspace{-1mm}
\section{Related work}
\label{others}
\vspace{-1mm}
\noindent\textbf{Single image deep motion deblurring.}
The task of deep motion deblurring for single images originated from global deblurring \cite{zhang2018learning,ren2021deblurring,tao2018scale,zamir2021multi,zhang2021deep,zhou2022lednet}. Pioneering deep global motion deblurring works utilize CNNs as basic layers and achieve promising improvements in image quality. Among them, DeepDeblur \cite{nah2017deep}, a multi-scale convolutional neural network, performs residual blocks to increase convergence speed. DeblurGAN \cite{kupyn2018deblurgan} and DeblurGAN-v2 \cite{kupyn2019deblurgan} introduce GANs and a perceptual loss to improve subjective quality. HINet \cite{chen2021hinet} applies Instance Normalization to boost performance. Recently, a CNN-based local motion deblurring method, LBAG \cite{li2022real}, bridges the gap between global and local motion deblurring by inserting gate modules at the end of MIMO-UNet architecture \cite{cho2021rethinking}. It predicts differentiable blur masks to reduce sharp backgrounds from modifications and guide the network to deblur locally.
Although the performance is significantly improved, CNN-based methods suffer from the content-independent interactions between images and convolution kernels, as well as the limitations of long-range dependency modeling.

Given the Vision Transformer's (ViT) \cite{dosovitskiy2020image} ability to capture long-range dependencies, its application to global deblurring tasks has seen a surge of interest. For example, Uformer \cite{wang2022uformer} employs window-based self-attention with a learnable multi-scale restoration modulator to capture both local and global dependencies. Restormer \cite{zamir2022restormer} utilizes multi-head attention and a feed-forward network to achieve long-range pixel interactions. In this paper, we build a Transformer-based local motion deblurring framework, LMD-ViT, that adaptively selects windows relevant to blurry regions for window-based self-attention and feed-forward operations, simultaneously benefiting from long-range modeling. 

\noindent\textbf{Vision Transformer acceleration.}
Transformers have proven valuable in deblurring tasks, yet their direct application in local motion deblurring for high-resolution images presents challenges concerning computational efficiency.
To solve the heavy computation problem of global self-attention in Transformers, researchers have presented several techniques. For example, Wang et al. adopted pyramid structures and spatial-reduction attention \cite{wang2021pyramid} in image classification, object detection, and segmentation tasks. Some methods partition image features into different windows and perform self-attention on local windows \cite{vaswani2021scaling, yang2021focal, wang2022uformer} for image restoration tasks. Some image classification methods gradually reduce tokens in processing by token-halting \cite{meng2022adavit,rao2021dynamicvit,yin2022vit} or token-merging \cite{liang2022not,bolya2022token}. Inspired by these advancements, we develop adaptive window pruning blocks (AdaWPT) to eliminate unnecessary tokens and focus deblurring only on blurred regions, which improves image quality and enables inference speed-up without compromising sharp regions.

\vspace{-1mm}
\section{Methodology}
\label{sec:method}
\vspace{-1mm}
\label{sec:method}
\begin{figure}[t]
    \centering
    \includegraphics[width = \textwidth,trim={0 0 10cm 0},clip]{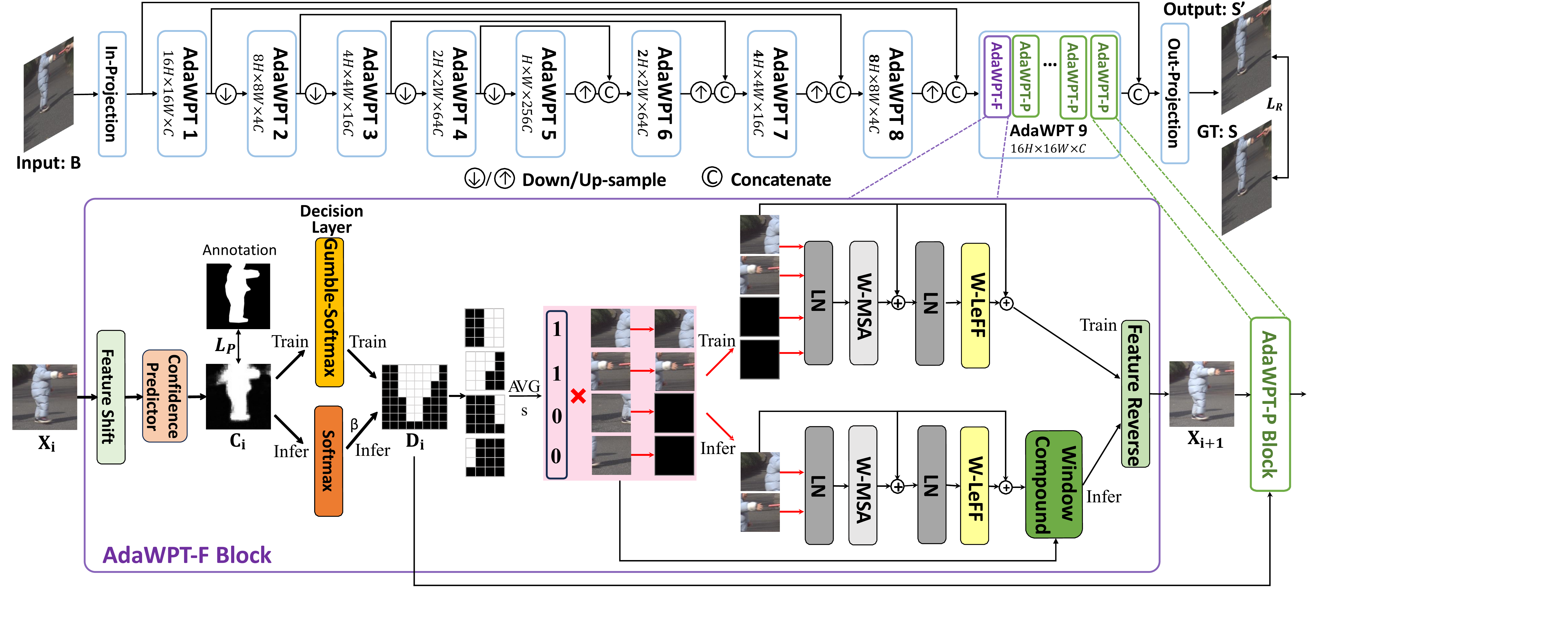}
    \caption{The architecture of LMD-ViT and AdaWPT block. LMD-ViT is built on a U-shape encoder-decoder structure with  AdaWPT blocks, which prune tokens at different resolutions. Each AdaWPT block contains an AdaWPT-F block and several AdaWPT-P blocks. AdaWPT-F predicts the blurriness confidence ($\mathbf{C_i}$) and the pruning decision ($\mathbf{D_i}$). AdaWPT-P follows the pruning decisions, neither producing the blurriness confidence nor calculating the pruning loss ($\mathcal{L_P}$). $\mathbf{X_i}$ and $\mathbf{X_{i+1}}$ denote the $\mathbf{i^{th}}$ and $\mathbf{{i+1}^{th}}$ feature map of each blocks, respectively. ``AVG'', ``$\beta$'', and ``s'' denote the average pooling operation, the pruning threshold in inference, and the threshold of the average pooling operation, respectively.}
    \vspace{-2mm}
    \label{fig:network_arch}
\end{figure}

\subsection{Model architecture}
\vspace{-1mm}
The architecture of our local motion deblurring vision Transformer (LMD-ViT) is shown at the top of \autoref{fig:network_arch}, which is a U-shaped network with an encoder stage, a bottleneck stage, and a decoder stage with skip connections. An in-projection/out-projection layer is placed at the beginning/end of the network to extract RGB images to feature maps or convert feature maps to RGB images.
The encoder, bottleneck, and decoder include a series of adaptive window-token pruning Transformer blocks (AdaWPT) and down-sampling/up-sampling layers. 
As a key component, AdaWPT removes local blurs by a window pruning strategy with a confidence predictor, a decision layer, and several Transformer layers. It is trained with a reconstruction loss ($\mathcal{L_R}$) and a pruning loss ($\mathcal{L_P}$) constrained by our carefully annotated blur masks. 
AdaWPT can be applied in any encoder/decoder/bottom-neck block and is flexible to prune at different resolutions.  As shown in \autoref{fig:motivation}, in our proposed LMD-ViT, windows are pruned coarsely in low-resolution blocks and finely in high-resolution blocks. This strikes a balance between computational complexity and accuracy.  
We detail the architectures and model hyper-parameters of LMD-ViT in Appendix \ref{sec:arch}. 

\subsection{Adaptive window pruning Transformer (AdaWPT)}
\vspace{-1mm}
As shown at the bottom of \autoref{fig:network_arch}, an adaptive window pruning Transformer block (AdaWPT) includes a First-AdaWPT block (AdaWPT-F) and several Post-AdaWPT blocks (AdaWPT-P). 
During training, each AdaWPT-F comprises a confidence predictor, a decision layer, a feature shift/reverse block, and several Transformer layers such as window-based multi-head self-attention (W-MSA) \citep{wang2022uformer}, window-based locally-enhanced feed-forward layer (W-LeFF), and layer normalization (LN). A window compound layer is added in inference. 
AdaWPT-F is responsible for predicting the confidence of blurriness by a confidence predictor, generating pruning decisions from a decision layer and pruning windows. In order to save computational resources, AdaWPT-P follows the decisions provided by AdaWPT-F to prune without producing the confidence of blurriness and decisions again. Therefore, AdaWPT-P does not contain a confidence predictor or a decision layer. In both kinds of AdaWPT, only the unpruned windows are fed into the window-based Transformer layers, significantly reducing computational costs and keeping sharp regions undistorted. Besides, a feature shift/reverse block is inserted before/after pruning to promote feature interactions.
We detail the modules in AdaWPT in the following paragraphs.
\vspace{-1mm}
\subsubsection{Confidence predictor}
\label{sec:cp}
\vspace{-1mm}
The confidence predictor predicts the confidence of blurriness for the input features $\mathbf{X_i}\in \mathbb{R}^n$. Tokens with higher confidence are more likely to be kept and others are removed.
Following \cite{rao2021dynamicvit}, the confidence predictor employs MLP layers to produce feature embeddings $e$ and predict the confidence map $\mathbf{C}$ using Softmax:
\begin{equation}
\small
        \mathbf{C_i} = \text{Softmax}(e_i),
        ~e_i = \text{Concat}\bigg(\text{MLP}\big(\text{MLP}(\mathbf{X_i}), \frac{\sum_{j=n}^{N}\mathbf{D}_j\cdot \text{MLP}(\mathbf{X_i}_j)}
        {\sum_{j=n}^{N}\mathbf{D}_j}\big)\bigg),
        i\in\mathbb{N_+},
    \label{predictor}
\end{equation}
where $\mathbf{D}$, initialized using $\mathbf{I}$, is the one-hot decision map to prune windows, which will be introduced in Section \ref{sec:prune}. $i$ denotes the $i^{th}$ AdaWPT block.
The confidence predictor is trained using an end-to-end reconstruction loss (introduced in Section \ref{sec:loss}) with Gumbel-Softmax parameterization (see Section \ref{sec:prune}), together with a pruning loss (introduced in Section \ref{sec:loss}) guided by our blurry mask annotations (introduced in Section \ref{sec:anno}).

\subsubsection{Decision layer}
\label{sec:prune}
The decision layer samples from the blurriness confidence map to generate binary pruning decisions $\mathbf{D}$, in which ``1'' represents tokens to be kept while ``0'' represents tokens to be abandoned. Although our goal is to perform window pruning, it is not practical to remove the zero tokens directly, for the absence of tokens halts backward propagation, and the different removal instances make parallel computing impossible in end-to-end training. To overcome this issue, we design the decision layer for training and testing, respectively. 

In training, we apply the Gumbel-Softmax re-parameterization \citep{jang2016categorical} as the decision layer, since it assures the gradients to flow through the network when sampling the training decision map $\mathbf{D^{tr}}$ from the training confidence map $\mathbf{C^{tr}}$:
\begin{equation}\small
   \mathbf{D^{tr}_i}(x,y) = \text{Gumbel-Softmax}(\mathbf{C^{tr}}(x,y)),
	\label{equ_decision}
\end{equation}
where (x,y) represents the coordinates of each window. 

In testing, we apply Softmax with a constant threshold $\beta$ as the decision layer: 
\begin{equation}\small
   \mathbf{D^{te}_i}(x,y) =
   \begin{cases}
       0,& \text{H}\big(\text{Softmax}\big(\mathbf{C^{te}}_i(x,y))\big)\big)< \beta\\
       1,& \text{H}\big(\text{Softmax}\big(\mathbf{C^{te}}_i(x,y))\big)\big)\geq \beta,
       \label{eq:softmax}
   \end{cases}
\end{equation}
where \text{H} functions to sparse the output of \text{Softmax} to $1$/$0$ when it is larger/fewer than $\beta$. The abandoned windows, that is, windows irrelevant to local blurs, are further set to zero by 
\begin{equation}
\small
    \mathbf{X_i'} = \mathbf{D_i}\cdot \mathbf{X_i}.
    \label{equa:decision}
\end{equation}
\subsubsection{Efficient Transformer layers with window pruning strategy}
Considering the quadratic computation costs with respect to a large number of tokens of high-resolution images, we employ Transformer layers in non-overlapping windows rather than in a global manner to accelerate training and inference. 
Specifically, we choose W-MSA layer \citep{wang2022uformer} for the self-attention (SA) operation. 
For the feed-forward structure, we develop a window-based locally-enhanced feed-forward layer (W-LeFF) as an alternative to LeFF \citep{wang2022uformer} which modifies global tokens. W-LeFF uses a $3\times3$ convolution with stride 1 and reflected padding 1 in independent windows. The reflected padding ensures the continuity at the window borders and obtains almost the same performance as LeFF \citep{wang2022uformer} (we compare the performance of LeFF and W-LeFF in Appendix \ref{sec:discus_wleff}). 

To enable parallel computing, we regard each window as a token group and prune based on windows. Each window owns a 0/1 decision, which is calculated by average pooling the decision map with a threshold $s$. Windows with an average $\geq s$ are regarded as the kept windows and the others as the abandoned windows. To promote content interactions among non-overlapping windows, we also apply relative position encoding \citep{wang2022uformer} in the attention module and shift/reverse the windows by half the window size at the beginning/end of the AdaWTP.

To accomplish both differentiable training and fast inference, we propose different pruning strategies with W-MSA and W-LeFF in training and testing, respectively. 
In training, to ensure back-propagation, all the windows including the abandoned windows go through W-MSA, W-LeFF, and LN sequentially to generate training features $\mathbf{X^t_{i+1}}$:
\begin{equation}\small
     \mathbf{{X^{tr}_{i+1}}'} = 
    \text{W-MSA}(\text{LN}(\mathbf{{X^{tr}_i}'})) + \mathbf{{X^{tr}_i}'},~
    \mathbf{X^{tr}_{i+1}} = 
    \text{W-LeFF}(\text{LN}(\mathbf{{X^{tr}_{i+1}}'})) + \mathbf{{X^{tr}_{i+1}}'}.
    \label{Train-SA-FFN}
\end{equation}
In testing, only the kept windows are processed by Transformer layers, which release a great number of unnecessary tokens to deblur. To enable future stages to perform global operations, we mend the abandoned windows to their original locations to compound a complete testing feature map $\mathbf{X^{te}_{i+1}}$ :
\begin{equation}\small
    \mathbf{{X^{te}_{i+1}}'} = 
    \text{W-MSA}(\text{LN}(\mathbf{{X^{te}_i}'}\geq s)) + \mathbf{{X^{te}_i}',~
    \mathbf{X^{te}_{i+1}} = 
    \text{W-LeFF}(\text{LN}(\mathbf{{X^{te}_{i+1}}'\geq s}})) + \mathbf{{X^{te}_{i+1}}'},
    \label{Infer-SA-FFN}
\end{equation}
where $s$ denotes the threshold in the average pooling operation (AVG).
\vspace{-1mm}
\subsection{Blur region annotation}
\label{sec:anno}
\vspace{-1mm}
To obtain the ground-truth local blur mask for supervised training of the confidence predictor, we carefully annotate the binary local blur masks of the ReLoBlur dataset \citep{li2022real}. We mark the blurred regions with pixel value $1$ and others with $0$, as shown in \autoref{fig:motivation}. To simplify the annotation process, we firstly select moving objects by the \textit{EISeg} software\footnote{\textit{EISeg} is provided by \url{https://github.com/PaddlePaddle/PaddleSeg/tree/release/2.7/EISeg}}, which can automatically segment different components within each object. Then, we manually refine the blurry regions by 1) expanding blurry components that were not included in the initial annotation; and 2) removing redundant sharp regions.
\subsection{Loss functions}
\label{sec:loss}
To guide adaptive window pruning and locally deblurring, we propose a pruning loss and combine it with a weighted reconstruction loss to form the total loss:
\begin{equation}\small
    \mathcal{L} = \mathcal{L_P} + \mathcal{L_R}.
\end{equation}
\noindent\textbf{Pruning loss.} We propose a pruning loss to constrain the blurriness confidence prediction:
\begin{equation}\small
    \mathcal{L_P} = \lambda_0 \sum_{i=1}^{n}\text{Cross-Entropy}(\mathbf{C_i},\text{Down-Sample}^i(\mathbf{M})),
\end{equation}
where $\lambda_0$=$0.01$, and the blur mask $\mathbf{M}$ is down-sampled to match the resolution of the confidence maps, and $i$ indicates a confidence predictor's location.

\noindent\textbf{Reconstruction loss.} To focus training on local blurred regions, following \cite{li2022real}, we apply weights $w$ on the reconstruction loss, which is a combination of $L_1$ loss, SSIM loss, and FFT loss: 
\begin{equation}\small
\begin{aligned}
    \mathcal{L_R} = w\mathcal{L_R'}(\mathbf{M}\cdot\mathbf{S'}, \mathbf{M}\cdot\mathbf{S})
                &+ (1-w)\mathcal{L_R'}(\mathbf{(1-M)}\cdot\mathbf{S'}, \mathbf{(1-M)}\cdot\mathbf{S}), \\
    \mathcal{L_R'} = \text{$L_1$}(\mathbf{S'}, \mathbf{S}) &+ \lambda_2 \text{$SSIM$}(\mathbf{S'}, \mathbf{S}) + \lambda_3 \text{$FFT$}(\mathbf{S'}, \mathbf{S}), \\ 
    \label{eq:loss}
\end{aligned}
\end{equation}
where $S$, and $S'$ denote the sharp ground truth, and the deblurred output. $w=0.8$, $\lambda_1$=$\lambda_2$=$1.0$, and $\lambda_3$=$0.1$ denote the weights of the loss functions.

\vspace{-1mm}
\begin{figure}[t]
    \centering
    \includegraphics[width = \textwidth]{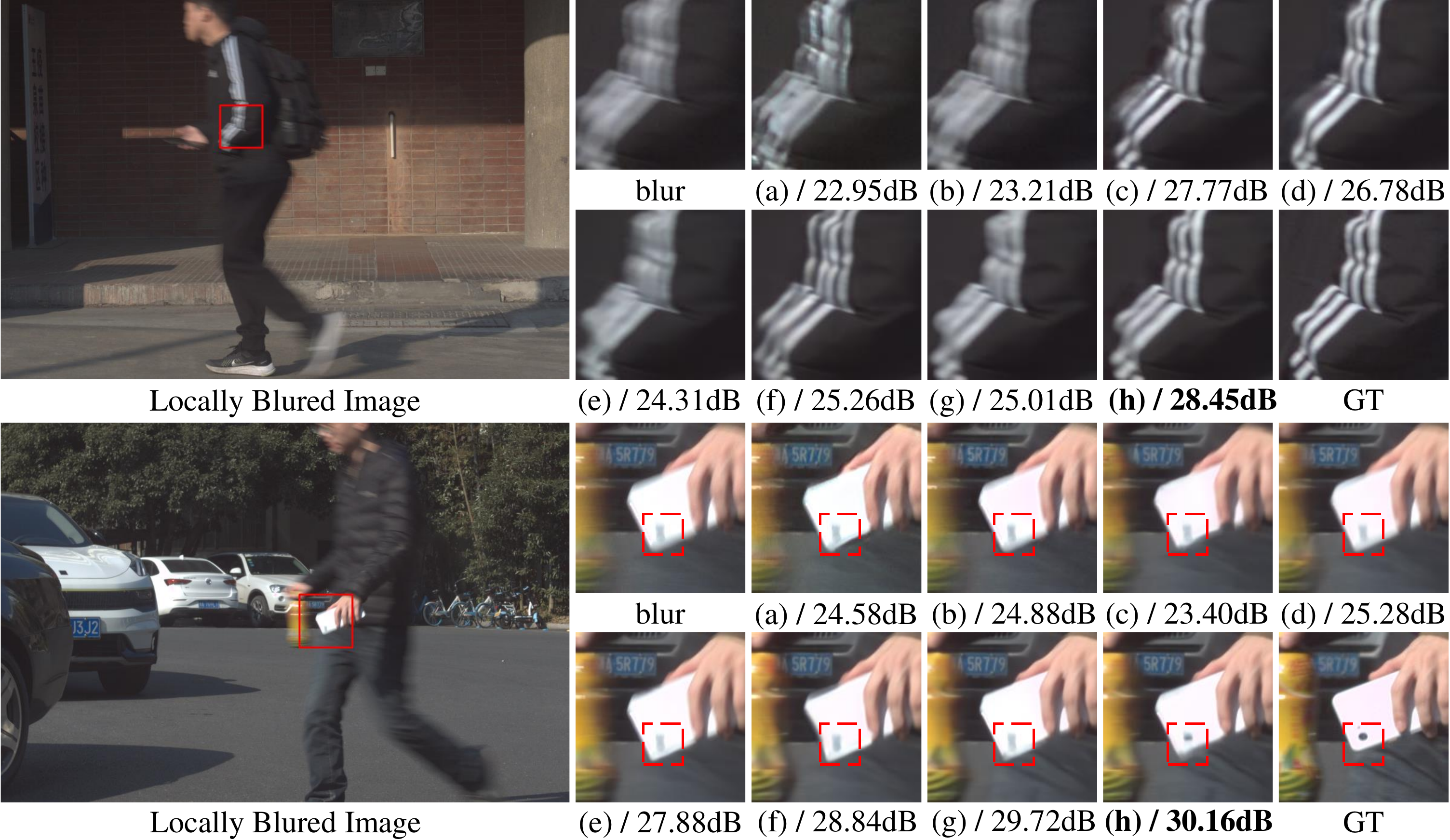}
    \caption{Visual comparisons of state-of-the-art methods for local motion deblurring. (a) DeepDeblur \citep{nah2017deep}; (b) DeblurGAN\_v2 \citep{kupyn2019deblurgan}; (c) HINet \citep{chen2021hinet}; (d) MIMO-UNet \citep{cho2021rethinking}; (e) LBAG \citep{li2022real}; (f) Restormer \citep{zamir2022restormer}; (g) Uformer \citep{wang2022uformer}; (h) LMD-ViT (ours).}
    \label{fig:main_deblurred}
\end{figure}
\vspace{-2mm}
\begin{figure}[t]
    \centering
    \includegraphics[width = \textwidth,trim={0cm 0cm 0cm 4cm}, clip=true]{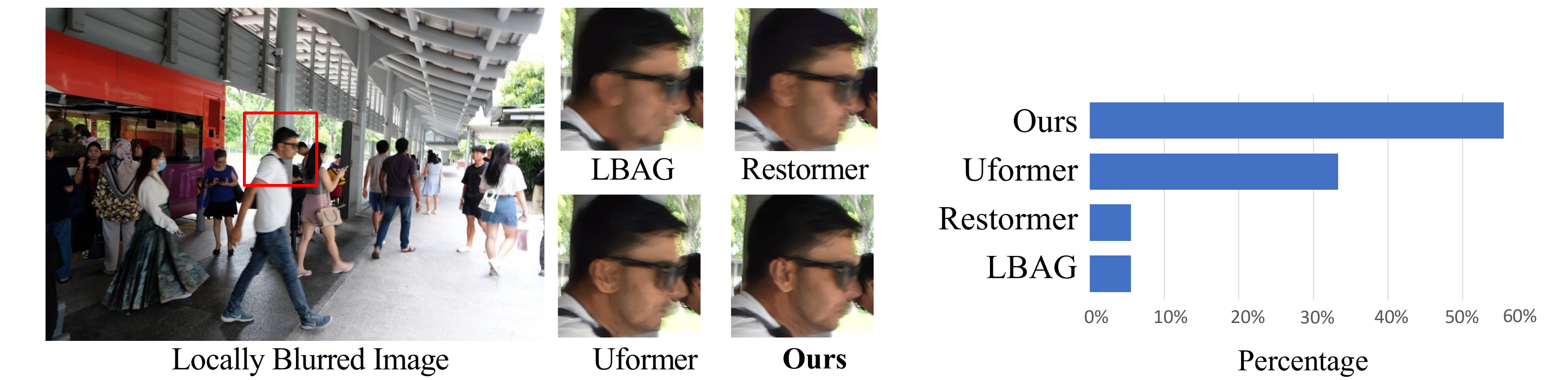}
    \caption{Results of user study for real-world local motion deblurring. The horizontal axis indicates the percentage of favoring each method.}
    \label{fig:userstudy}
\end{figure}

\section{Experiments and analyses}
\vspace{-1mm}
\subsection{Experimental settings}
We train LMD-ViT using AdamW optimizer \citep{kingma2014adam} with the momentum terms of (0.9, 0.999), a batch size of 12, and an initial learning rate of $2\times10^{-4}$ that is updated every 2k steps by a cosine annealing schedule \citep{loshchilov2016sgdr}. We set the window size of AdaWPT to $8\times8$, and the initial embedded dim to 32 which is doubled after passing each down-sampling layer. 
LMD-ViT is trained on the GoPro dataset \citep{nah2017deep} and ReLoBlur dataset \citep{li2022real} together, in order to enable both local and global motion deblurring. The sampling ratio of the GoPro training data~\citep{nah2017deep} and the ReLoBlur training data~\citep{li2022real} is set close to 1:1.

For a fair comparison, we train the baseline methods using the same datasets and cropping strategy. The model configurations of the compared deblurring methods follow their origin settings.

We evaluate our proposed LMD-ViT and baseline methods on the ReLoBlur testing dataset \citep{li2022real} with the full image size of 2152$\times$1436 on 1 Nvidia A100 GPU. In addition to the commonly-used PSNR and SSIM \citep{wang2004image} metrics, we follow the approach of \cite{li2022real} and calculate weighted PSNR and weighted SSIM specifically for the blurred regions. This allows us to better assess the local deblurring performance. We provide the evaluation results in the following sections and Appendix \ref{sec:vis_results}. To measure model efficiency, we report the inference time, FLOPs, and model parameters.
\vspace{-1mm}
\subsection{Experimental results}
\vspace{-1mm}
\noindent\textbf{Evaluations on public datasets.}
We first compare the proposed LMD-ViT with both CNN-based methods \citep{nah2017deep,kupyn2019deblurgan,chen2021hinet,cho2021rethinking,li2022real} and Transformer-based methods \citep{zamir2022restormer,wang2022uformer} on the ReLoBlur dataset \citep{li2022real} for local motion deblurring. 
As depicted in \autoref{fig:motivation} and \autoref{fig:main_deblurred}, LMD-ViT exhibits superior performance compared to other state-of-the-art methods, producing clearer outputs with enhanced details. Notably, the white stripes on the student's suit and the mobile phone show significant blur reduction without artifacts and closely resemble the ground truth. We notice that the number of preserved windows slightly exceeds the number of windows in the annotated blurry areas. This is to ensure that the preserved windows effectively cover as much of the blurry area as possible.
Quantitative evaluation results are presented in \autoref{table:resutls}, which bolds the best local deblurring performance. Compared with CNN-based methods, our proposed LMD-ViT achieves an improvement of 0.50 dB in PSNR and 0.95 dB in weighted PSNR, while maintaining a comparable or even faster inference speed. Compared with Transformer-based methods, LMD-ViT demonstrates significant reductions (-66\%) in FLOPs and inference time without sacrificing performance (PSNR +0.28dB), thanks to our adaptive window pruning modules. The PSNR and SSIM scores of LMD-ViT also exceed other Transformer-based deblurring methods, because our proposed pruning strategy prohibits the network from destroying sharp regions. However, global deblurring methods (e.g., Uformer~\citep{wang2022uformer} and  Restormer~\citep{zamir2022restormer}) treat every region equally and may distort or blur the sharp regions inevitably.
\begin{table}[t]\small
\centering
\caption{Quantitative comparisons on the local deblurring dataset. ``PSNR$_w$'', ``SSIM$_w$'', ``Time'' and ``Params'' denote weighted PSNR, weighted SSIM, inference time, and model parameters respectively. We bold the best result under each evaluation metric.}
\label{table:resutls}
\setlength{\tabcolsep}{0.5mm}
\begin{tabular}{c| c |c c c c |c c c}
\toprule
Categories & Methods  & $\uparrow$PSNR & $\uparrow$SSIM  & $\uparrow$PSNR$_w$ & $\uparrow$SSIM$_w$ & Time & Params &FLOPs\\
\midrule
& DeepDeblur~\citep{nah2017deep}      & 33.02 & 0.8904 & 28.29 & 0.8398 & 0.50s & 11.72M & 17.133T\\
& DeblurGAN-v2~\citep{kupyn2019deblurgan}     & 33.26 & 0.8975 & 28.29 & 0.8489  & 0.07s & 5.076M & 0.989T\\
CNNs & HINet~\citep{chen2021hinet}  & 34.40
 & 0.9160 & 28.82 & 0.8672  & 0.31s & 88.67M & 8.696T\\
& MIMO-UNet~\citep{cho2021rethinking}        & 34.64 & 0.9247 & 29.17 & 0.8766
  &  0.51s & 16.11M & 7.850T\\
& LBAG~\citep{li2022real}  & 34.92 & 0.9318
 & 29.30 & 0.8946  & 0.51s & 16.11M &7.852T \\
\midrule
& Restormer~\citep{zamir2022restormer}  & 34.85  & 0.9271 & 29.28 & 0.8785  & 3.72s &26.13M & 6.741T\\
Transformers & Uformer-B~\citep{wang2022uformer}  & 35.14 & 0.9277 & 29.92 & 0.8865  & 1.31s & 50.88M &4.375T\\
& LMD-ViT  & \textbf{35.42} & \textbf{0.9285} & \textbf{30.25}  & \textbf{0.8938} & 0.56s & 54.50M &1.485T\\
\bottomrule
\end{tabular}
\vspace{-2mm}
\end{table}
\begin{figure}[t]
    \vspace{-2mm}
    \centering
    \includegraphics[width = \textwidth]{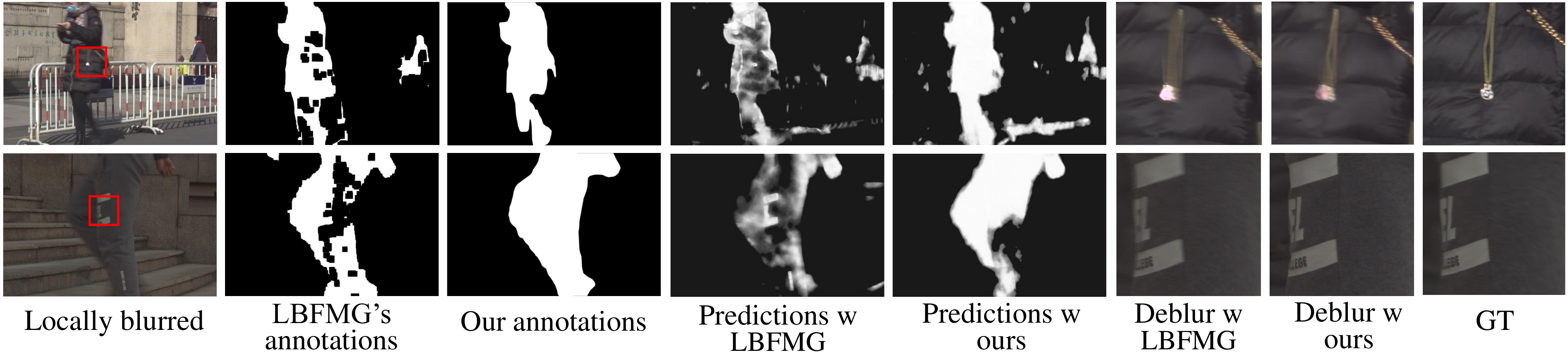}
    \caption{Comparisons of blur mask prediction. With more accurate annotation, our method produces better blur masks than that of LBFMG~\citep{li2022real}, benefiting the deblurring performance.}
    \label{fig:annotate}
\end{figure}
\begin{figure}[t]
    \centering
    \vspace{-2mm}
    \includegraphics[width = \textwidth]{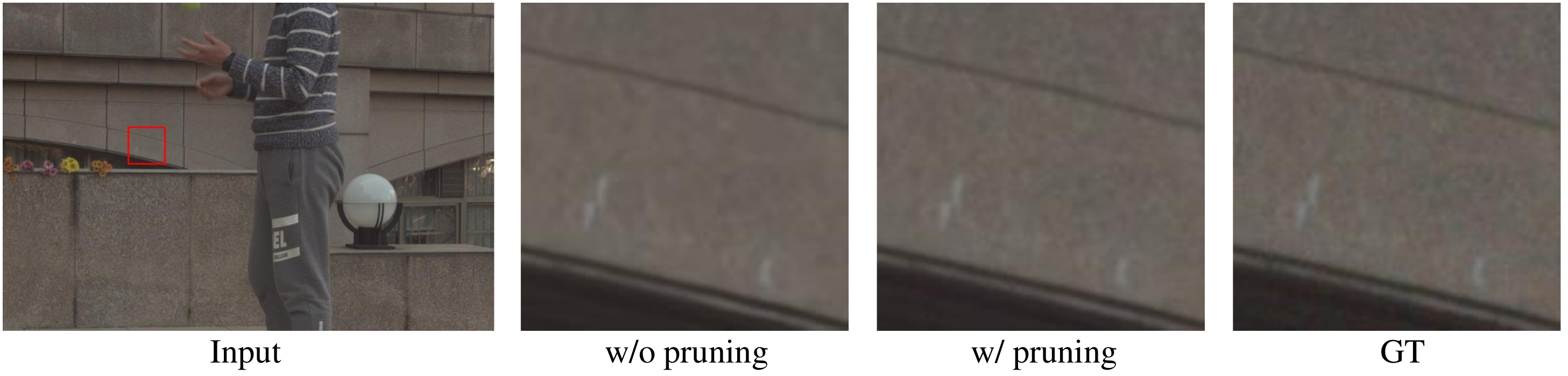}
    \caption{Evaluations for sharp region quality preservation with or without the pruning strategy. Without pruning, the network modifies global pixels and blurs the sharp backgrounds. In contrast, with pruning, sharp backgrounds are preserved well.}
    \label{fig:prune}
    \vspace{-2mm}
\end{figure}
Additionally, our proposed LMD-ViT could deblur globally. We evaluate LMD-ViT's global deblurring performance on the GoPro testing dataset \citep{nah2017deep}. 
Our proposed LMD-ViT obtains comparable deblurring performance to global deblurring methods. We also observed that with a globally blurry input, LMD-ViT does not prune windows because the confidence of blurriness of all regions is nearly 100\%. We explain the global deblurring performances in detail in Appendix \ref{sec:global}.
\noindent\textbf{User study on real-world photos.} To validate the effectiveness of our proposed model in real-world locally blurred images, we capture 18 locally blurred RGB images with a resolution of 6000$\times$4000. We conduct a comparison between our method and the top 3 methods listed in \autoref{table:resutls} on these images. Since the ground truths for these blurred images are not available, we conducted a user study involving 30 participants who are passionate about photography. Each participant is presented with randomly selected deblurred images and asked to choose the most visually promising deblurred image. As shown in \autoref{fig:userstudy}, our proposed method exhibits robust performance on real-world locally blurred images, with sharper reconstructed edges and consistent content. Moreover, it is the most preferred method among the participants when compared to other approaches.
\vspace{-2mm}
\subsection{Analyses}
\vspace{-1mm}
To further analyze the ability of our proposed method, we analyze the effectiveness of the window pruning strategy and blur mask annotations in this section, respectively. Due to the limited space, the analyses of W-LeFF and the number of feature channels are illustrated in Appendix \ref{sec:discus}.
\begin{table}[t]\small
\centering
\caption{The effectiveness of window pruning. ``PSNR$_w$'', ``SSIM$_w$'', ``Time'', and ``Params'' denote weighted PSNR, weighted SSIM, inference time, and model parameters, respectively.} 
\label{table:prune_ablation}
\setlength{\tabcolsep}{2.2mm}
\begin{tabular}{c |c |c | c c c c | c c c}
\toprule
No. & Pruning block & $\beta$  & $\uparrow$PSNR & $\uparrow$SSIM & $\uparrow$PSNR$_w$ & $\uparrow$SSIM$_w$ & Time & Params &FLOPs\\
\midrule
1 & AdaWPT 1$\sim$9  & 0.5 & 35.42  & 0.9289 & 30.25 & 0.8938   &  0.56s &54.50M & 1.485T\\
\midrule
2 & AdaWPT 2$\sim$8  & 0.5 & 35.41  & 0.9289  & 30.34  & 0.8928  &  0.70s & 53.94M &1.592T\\
3 & AdaWPT 3$\sim$7  & 0.5 & 35.44  & 0.9290  & 30.38  & 0.8936  &  0.77s &53.49M  &2.021T\\
4 & AdaWPT 4$\sim$6  & 0.5 & 35.37  & 0.9293  & 30.39  & 0.8931  &  1.07s &53.06M  &3.105T\\
5 & None     & 0.5 & 35.36  & 0.9280 & 30.22 & 0.8930    & 1.30s & 50.39M &4.376T\\
6 & AdaWPT 1$\sim$9  & 0.2 & 35.36  &  0.9289 & 30.21 &0.8931    &  0.95s & 54.50M & 1.911T \\
7 & AdaWPT 1$\sim$9  & 0.3 & 35.41  & 0.9291 &  30.28 &0.8934    &  0.80s & 54.50M &1.671T \\
8 & AdaWPT 1$\sim$9  & 0.4 &35.43  &  0.9290 & 30.28 & 0.8935    &  0.69s & 54.50M & 1.556T\\
9 & AdaWPT 1$\sim$9  & 0.6 & 35.35  & 0.9284 & 30.18 & 0.8922   &  0.49s & 54.50M &1.405T\\
10 & AdaWPT 1$\sim$9  & 0.7 & 35.32  & 0.9281 & 30.14 & 0.8918    &  0.43s & 54.50M &1.327T\\
\bottomrule
\end{tabular}
\vspace{-2mm}
\end{table}
\begin{table}[t]\small
\centering
\caption{The effectiveness of our blur mask annotation. ``PSNR$_w$'' and ``SSIM$_w$''  denote weighted PSNR, and weighted SSIM, respectively.}
\label{table:annotate_resutls}
\setlength{\tabcolsep}{1mm}
\begin{tabular}{c |c |c c c c c}
\toprule
No. & Methods & $\uparrow$PSNR & $\uparrow$SSIM & $\uparrow$PSNR$_w$ & $\uparrow$SSIM$_w$ & $\uparrow$Precision \\
\midrule
1 & LBAG \citep{li2022real} (w/ LBFMG \citep{li2022real}) & 34.83 & 0.9264 & 28.31 & 0.8711   & 0.6632 \\
2 & LBAG \citep{li2022real} (w/ ours)   & 34.92 & 0.9318 & 29.30 & 0.8946   & 0.6949 \\
3 & LMD-ViT (w/ LBFMG \citep{li2022real}) & 35.31 & 0.9270 & 30.14 & 0.8911  & 0.8611 \\
4 & LMD-ViT (w/ ours) & 35.42 & 0.9289 & 30.25 & 0.8938  & 0.9680 \\
\bottomrule
\end{tabular}
\vspace{-2mm}
\end{table}

\subsubsection{Window pruning strategy}
We analyze the effects of our window pruning strategy in three aspects: the number of pruning blocks, the pruning threshold $\beta$ during inference, and pruning precision. 

\noindent{\textbf{The number of pruning blocks.}} We fix the pruning threshold $\beta=0.5$ and change the number of pruning blocks. For blocks without pruning, we force the decision maps $\mathbf{D}$ in Equation \ref{equa:decision} to be the all-one matrix. From line 1 to line 5 of \autoref{table:prune_ablation}, we find that pruning more blocks results in fewer model parameters, a higher inference speed, and more dropping scores. Notably, although we apply a window pruning mechanism in all 9 blocks, the scores outperform other baseline models listed in \autoref{table:resutls}. Additionally, we conduct visual comparisons of all-pruned architecture (line 1) with non-pruning architecture (line 5), as shown in \autoref{fig:prune}. In the all-pruned network, most of the sharp regions are pruned and are not destroyed. In contrast, the sharp regions turn blurry or distorted when processed by the non-pruned network. This indicates that our adaptive window pruning mechanism can prevent the sharp regions from being destroyed. The non-pruning network treats every region equally like global deblurring networks (e.g., MIMO-UNet, \citep{cho2021rethinking}, Uformer \citep{wang2022uformer} and Restormer \citep{zamir2022restormer}) that may harm the sharp regions inadvertently. 

\noindent{\textbf{Pruning threshold $\beta$}.} We fix the pruning blocks and adjust the pruning threshold $\beta$ from 0.2 to 0.7 with 0.1 as the interval. Comparing line 1, lines 6 to line 10 in \autoref{table:prune_ablation}, we find that the testing performance slightly varies with different pruning thresholds $\beta$, resulting in different confidence levels and decision boundaries. Inferring with a lower nor a higher pruning threshold is neither reasonable, because the former makes the confidence predictor more inclusive, which potentially leads to fewer sharp windows to be pruned and a slower inference speed, while the latter makes the confidence predictor more conservative, which leads to faster inference speed but filters out the necessary blurry windows. To achieve a balance, we choose $\beta=0.5$ as it obtains relatively high evaluation scores and fast inference speed.

\noindent\textbf{Pruning precision}

Our proposed adaptive window pruning Transformer block (AdaWPT) enables blur region selection as well as removing sharp windows. From the first pruning block (AdaWPT 1) to the last pruning block (AdaWPT 9), the preserved patches gradually correspond to the ground-truth blur annotations, as shown in \autoref{fig:motivation}. To verify whether blurred regions are mistakenly pruned, we calculate the precision of window pruning at the last pruning block: $Precision = TP/(TP + FP)$, where $TP$ refers to the number of pixels that our model correctly predicts to be blurry and preserved to process.  $FP$ refers to the number of pixels that our model incorrectly identifies to be blurry but are actually sharp, which are also pixels that our model accidentally preserved but should be actually removed. The average precision per block in the ReLoBlur testing dataset \citep{li2022real} is 96.8\%, suggesting that the preserved windows cover most blurred regions and AdaWPT significantly prunes sharp regions.

\subsubsection{Blur mask annotation}
To verify the effectiveness of our blur mask annotations, we conduct experiments on a local deblurring method LBAG \citep{li2022real} and our proposed LMD-ViT, as illustrated in \autoref{table:annotate_resutls}. For LBAG \citep{li2022real} which predicts blur masks as gates for local deblurring, our manually annotated blur masks enhance both image similarity and blur detection precision, comparing line 1 and line 2. \autoref{fig:annotate} also shows that the LBAG blur detection restrained by LBFMG masks \citep{li2022real} leads to more blur detection errors. When testing on LMD-ViT, our blur mask annotations improve the evaluation metrics obviously. 
However, performances of LMD-ViT drops when supervised by the binary masks generated by LBFMG \citep{li2022real}. Because the LBFMG annotations \citep{li2022real} contain holes in blurred regions and noise in sharp regions, which may confuse AdaWTP to select blurry tokens. 
 
\section{Conclusion}
In this paper, we presented an adaptive and efficient approach, LMD-ViT, the first sparse vision Transformer for restoring high-resolution images affected by local motion blurs. LMD-ViT is built upon our novel adaptive window pruning Transformer blocks (AdaWPT), which utilize blur-aware confidence predictors to estimate the level of blur confidence in the feature domain. This information is then used to adaptively prune unnecessary windows in low-confidence regions. To train the confidence predictor, we designed an end-to-end reconstruction loss with Gumbel-Softmax re-parameterization, along with a pruning loss guided by our meticulously annotated blur masks. Extensive experiments demonstrate that our method effectively eliminates local motion blur while ensuring minimal deformation of sharp regions, resulting in a significant improvement in image quality and inference speed. Due to limited page space, we discuss the limitations and broader impacts in Appendix \ref{sec:lim_imp}.
\bibliography{neurips_2023}

\begin{thebibliography}{10}

\bibitem{bolya2022token}
Daniel Bolya, Cheng-Yang Fu, Xiaoliang Dai, Peizhao Zhang, Christoph Feichtenhofer, and Judy Hoffman.
\newblock {Token Merging}: Your vit but faster.
\newblock In {\em ICLR}, 2022.

\bibitem{chen2021hinet}
Liangyu Chen, Xin Lu, Jie Zhang, Xiaojie Chu, and Chengpeng Chen.
\newblock {HINet}: Half instance normalization network for image restoration.
\newblock In {\em CVPR}, 2021.

\bibitem{cho2021rethinking}
Sung-Jin Cho, Seo-Won Ji, Jun-Pyo Hong, Seung-Won Jung, and Sung-Jea Ko.
\newblock Rethinking coarse-to-fine approach in single image deblurring.
\newblock In {\em ICCV}, 2021.

\bibitem{dosovitskiy2020image}
Alexey Dosovitskiy, Lucas Beyer, Alexander Kolesnikov, Dirk Weissenborn, Xiaohua Zhai, Thomas Unterthiner, Mostafa Dehghani, Matthias Minderer, Georg Heigold, Sylvain Gelly, et~al.
\newblock An image is worth 16x16 words: Transformers for image recognition at scale.
\newblock In {\em ICLR}, 2021.

\bibitem{guo2023shadowformer}
Lanqing Guo, Siyu Huang, Ding Liu, Hao Cheng, and Bihan Wen.
\newblock Shadowformer: Global context helps image shadow removal.
\newblock In {\em AAAI}, 2023.

\bibitem{jang2016categorical}
Eric Jang, Shixiang Gu, and Ben Poole.
\newblock Categorical reparameterization with gumbel-softmax.
\newblock In {\em ICLR}, 2017.

\bibitem{kingma2014adam}
Diederik~P Kingma and Jimmy Ba.
\newblock Adam: A method for stochastic optimization.
\newblock In {\em ICLR}, 2015.

\bibitem{kupyn2018deblurgan}
Orest Kupyn, Volodymyr Budzan, Mykola Mykhailych, Dmytro Mishkin, and Ji{\v{r}}{\'\i} Matas.
\newblock Deblurgan: Blind motion deblurring using conditional adversarial networks.
\newblock In {\em CVPR}, 2018.

\bibitem{kupyn2019deblurgan}
Orest Kupyn, Tetiana Martyniuk, Junru Wu, and Zhangyang Wang.
\newblock {DeblurGAN-v2}: Deblurring (orders-of-magnitude) faster and better.
\newblock In {\em ICCV}, 2019.

\bibitem{li2022real}
Haoying Li, Ziran Zhang, Tingting Jiang, Peng Luo, and Huajun Feng.
\newblock Real-world deep local motion deblurring.
\newblock In {\em AAAI}, 2023.

\bibitem{liang2021swinir}
Jingyun Liang, Jiezhang Cao, Guolei Sun, Kai Zhang, Luc Van~Gool, and Radu Timofte.
\newblock {SwinIR}: Image restoration using swin transformer.
\newblock In {\em ICCVW}, 2021.

\bibitem{liang2022not}
Youwei Liang, Chongjian Ge, Zhan Tong, Yibing Song, Jue Wang, and Pengtao Xie.
\newblock Not all patches are what you need: Expediting vision transformers via token reorganizations.
\newblock In {\em ICLR}, 2022.

\bibitem{liu2021swin}
Ze~Liu, Yutong Lin, Yue Cao, Han Hu, Yixuan Wei, Zheng Zhang, Stephen Lin, and Baining Guo.
\newblock {Swin Transformer}: Hierarchical vision transformer using shifted windows.
\newblock In {\em ICCV}, 2021.

\bibitem{loshchilov2016sgdr}
Ilya Loshchilov and Frank Hutter.
\newblock Sgdr: Stochastic gradient descent with warm restarts.
\newblock In {\em ICLR}, 2017.

\bibitem{meng2022adavit}
Lingchen Meng, Hengduo Li, Bor-Chun Chen, Shiyi Lan, Zuxuan Wu, Yu-Gang Jiang, and Ser-Nam Lim.
\newblock {AdaViT}: Adaptive vision transformers for efficient image recognition.
\newblock In {\em CVPR}, 2022.

\bibitem{nah2017deep}
Seungjun Nah, Tae Hyun~Kim, and Kyoung Mu~Lee.
\newblock Deep multi-scale convolutional neural network for dynamic scene deblurring.
\newblock In {\em CVPR}, 2017.

\bibitem{rao2021dynamicvit}
Yongming Rao, Wenliang Zhao, Benlin Liu, Jiwen Lu, Jie Zhou, and Cho-Jui Hsieh.
\newblock Dynamicvit: Efficient vision transformers with dynamic token sparsification.
\newblock In {\em NeurIPS}, 2021.

\bibitem{ren2021deblurring}
Wenqi Ren, Jiawei Zhang, Jinshan Pan, Sifei Liu, Jimmy Ren, Junping Du, Xiaochun Cao, and Ming-Hsuan Yang.
\newblock Deblurring dynamic scenes via spatially varying recurrent neural networks.
\newblock {\em TPAMI}, 2021.

\bibitem{schelten2014localized}
Kevin Schelten and Stefan Roth.
\newblock Localized image blur removal through non-parametric kernel estimation.
\newblock In {\em ICCV}, 2014.

\bibitem{tao2018scale}
Xin Tao, Hongyun Gao, Xiaoyong Shen, Jue Wang, and Jiaya Jia.
\newblock Scale-recurrent network for deep image deblurring.
\newblock In {\em CVPR}, 2018.

\bibitem{vaswani2021scaling}
Ashish Vaswani, Prajit Ramachandran, Aravind Srinivas, Niki Parmar, Blake Hechtman, and Jonathon Shlens.
\newblock Scaling local self-attention for parameter efficient visual backbones.
\newblock In {\em CVPR}, 2021.

\bibitem{vaswani2017attention}
Ashish Vaswani, Noam Shazeer, Niki Parmar, Jakob Uszkoreit, Llion Jones, Aidan~N. Gomez, Lukasz Kaiser, and Illia Polosukhin.
\newblock Attention is all you need.
\newblock In {\em NeurIPS}, 2017.

\bibitem{wang2021pyramid}
Wenhai Wang, Enze Xie, Xiang Li, Deng-Ping Fan, Kaitao Song, Ding Liang, Tong Lu, Ping Luo, and Ling Shao.
\newblock Pyramid vision transformer: A versatile backbone for dense prediction without convolutions.
\newblock In {\em ICCV}, 2021.

\bibitem{wang2022uformer}
Zhendong Wang, Xiaodong Cun, Jianmin Bao, Wengang Zhou, Jianzhuang Liu, and Houqiang Li.
\newblock Uformer: A general u-shaped transformer for image restoration.
\newblock In {\em CVPR}, 2022.

\bibitem{wang2004image}
Zhou Wang, Alan~C Bovik, Hamid~R Sheikh, and Eero~P Simoncelli.
\newblock Image quality assessment: from error visibility to structural similarity.
\newblock {\em IEEE transactions on image processing}, 13(4):600--612, 2004.

\bibitem{yang2021focal}
Jianwei Yang, Chunyuan Li, Pengchuan Zhang, Xiyang Dai, Bin Xiao, Lu~Yuan, and Jianfeng Gao.
\newblock Focal self-attention for local-global interactions in vision transformers.
\newblock In {\em NeurIPS}, 2021.

\bibitem{yin2022vit}
Hongxu Yin, Arash Vahdat, Jose~M Alvarez, Arun Mallya, Jan Kautz, and Pavlo Molchanov.
\newblock {A-ViT}: Adaptive tokens for efficient vision transformer.
\newblock In {\em CVPR}, 2022.

\bibitem{zamir2022restormer}
Syed~Waqas Zamir, Aditya Arora, Salman Khan, Munawar Hayat, Fahad~Shahbaz Khan, and Ming-Hsuan Yang.
\newblock Restormer: Efficient transformer for high-resolution image restoration.
\newblock In {\em CVPR}, 2022.

\bibitem{zamir2021multi}
Syed~Waqas Zamir, Aditya Arora, Salman Khan, Munawar Hayat, Fahad~Shahbaz Khan, Ming-Hsuan Yang, and Ling Shao.
\newblock Multi-stage progressive image restoration.
\newblock In {\em CVPR}, 2021.

\bibitem{zhang2021deep}
Jiawei Zhang, Jinshan Pan, Daoye Wang, Shangchen Zhou, Xing Wei, Furong Zhao, Jianbo Liu, and Jimmy Ren.
\newblock Deep dynamic scene deblurring from optical flow.
\newblock {\em IEEE Transactions on Circuits and Systems for Video Technology}, 32(12):8250--8260, 2021.

\bibitem{zhang2018learning}
Shanghang Zhang, Xiaohui Shen, Zhe Lin, Radom{\'\i}r M{\v{e}}ch, Joao~P Costeira, and Jos{\'e}~MF Moura.
\newblock Learning to understand image blur.
\newblock In {\em CVPR}, 2018.

\bibitem{zhou2022lednet}
Shangchen Zhou, Chongyi Li, and Chen Change~Loy.
\newblock {LEDNet}: Joint low-light enhancement and deblurring in the dark.
\newblock In {\em ECCV}, 2022.

\end{thebibliography}
\bibliographystyle{plain}
\appendix
\section{LMD-ViT architecture and model hyper-parameters}
\label{sec:arch}
As shown in \autoref{fig:network_arch} of our main paper, our proposed LMD-ViT is a U-shaped network with one in-projection layer, four encoder stages, one bottleneck stage, four decoder stages, and one out-projection layer. Skip connections are set up between the encoder stage and the decoder stage. A locally blurred input image $\mathbf{B} \in \mathbb{R}$ with a shape
$H\times W\times3$ firstly goes through an in-projection block, which consists of a $3\times 3$ convolutional layer, a LeakyReLU layer, and a layer normalization block, to extract low-level features as a feature map $\mathbf{X}\in \mathbb{R}$. The feature map then passes four encoder stages, each of which includes a series of AdaWPT Transformer blocks and one down-sampling layer. AdaWPT uses a blur-aware confidence predictor and Gumble-Softmax re-parameterization to select blur-related tokens. Only the selected tokens are forwarded to Transformer layers including window-based self-attention (W-MSA), window-based locally-enhanced feed-forward layer (W-LeFF), and layer normalization (LN). The down-sampling layer down-samples the feature map size by 2 times and doubles the channels using $4\times 4$ convolution with stride 2. The feature map's shape turns to ${\frac{H}{2^i}\times\frac{W}{2^i}\times3}, i\in {\{1, 2, 3, 4\}}$ after $i$ encoder stages, and has the smallest resolution in the bottleneck, where it can sense the longest dependencies in two AdaWPTs. After the bottleneck stage, the feature map goes through four decoder stages, each of which owns an up-sampling layer and a series of AdaWTP blocks. The up-sampling layer uses $2\times 2$ transposed convolution with stride 2 to reduce half of the feature channels and double the size of the feature map. The features put into the AdaWPT blocks are concatenations of the up-sampled features and the corresponding features from the symmetrical encoder stages through skip connections. Finally, the feature map passes the out-projection block which reshapes the flattened features to 2D feature maps and applies a $3\times 3$ convolution layer to obtain a residual image $\mathbf{R}$. The restored sharp image $\mathbf{S'}$ is obtained by $\mathbf{S’}= \mathbf{B} + \mathbf{R}$.

\section{Global deblurring performances}
\label{sec:global}
\begin{figure}[t]
    \vspace{-1mm}
    \centering
    \includegraphics[width = \textwidth]{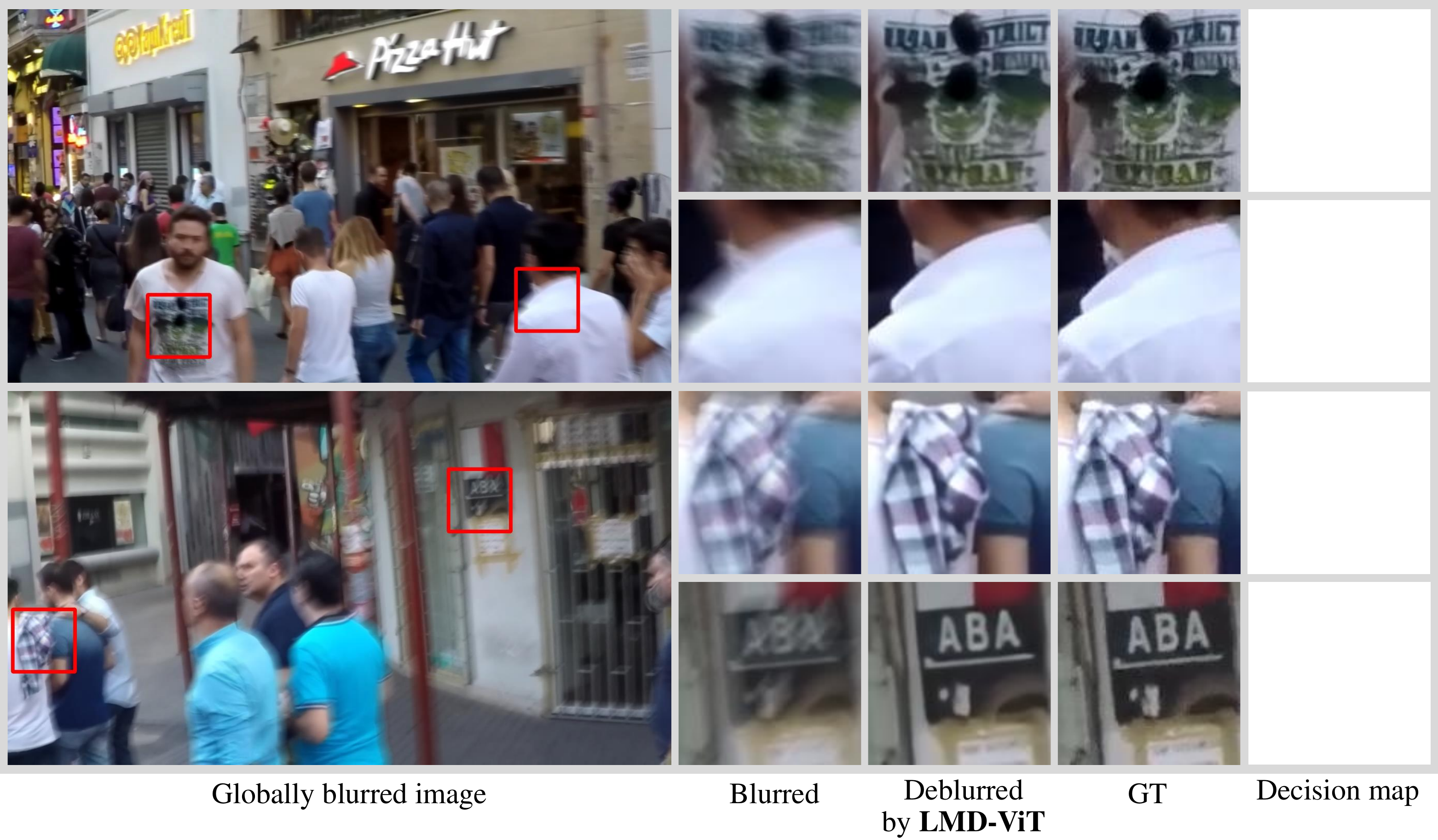}
    \caption{Global deblurring performance of our proposed LMD-ViT. The decision map denotes the pruning decisions. With a globally blurry input, the predicted blurriness confidence approaches 100\%. Therefore, our network preserves all windows and nearly all the values in decision maps equal to 1, which are represented as an all-white color.}
    \label{fig:sup_global}
\end{figure}

Our proposed LMD-ViT takes into account both local and global motion deblurring. To prove that our method could also deblur globally, we test our model on the GOPRO testing dataset \citep{nah2017deep}. With a globally blurry input, the predicted blurriness confidence approaches 100\%. Therefore, LMD-ViT preserves all windows, represented by the white color in the 5$^{th}$ column of \autoref{fig:sup_global}. We compare our proposed models with other state-of-the-art Transformer-based deblurring methods in Table \autoref{tab:quan_gopro}. The results show that our proposed method obtains comparable performance to the state-of-the-art global deblurring Transformers. The scores of our method are slightly lower than Uformer \citep{wang2022uformer} because our proposed LMD-ViT utilizes a window pruning strategy. The pruning precision (as described in Section ) is not 100\%. Some tokens miss the deblurring manipulations in some blocks and therefore they are not clear enough. The visual comparison on the GoPro testing dataset~\citep{nah2017deep} also shows that LMD-ViT is not inferior to other baseline methods. It can effectively eliminate global blurriness and restore sharp textures.
\begin{table}[h]\small
\centering
\caption{Quantitative comparisons on the global deblurring dataset. All the methods are trained with the ReLoBlur dataset and the Gopro dataset together.}
\label{tab:quan_gopro}
\begin{tabular}{c| c c c }
\toprule
Evalution metrics & Restormer~\citep{zamir2022restormer} & Uformer-B~\citep{wang2022uformer} &LMD-ViT (ours)\\
\midrule
PNSR      & 32.15 & 32.51 &32.16\\
SSIM      & 0.9305 & 0.9376 & 0.9318\\
\bottomrule
\end{tabular}
\vspace{-3mm}
\end{table}
\begin{figure}[t]
    \centering
    \includegraphics[width = \textwidth,trim={0cm 8cm 0cm 0cm}, clip=true]{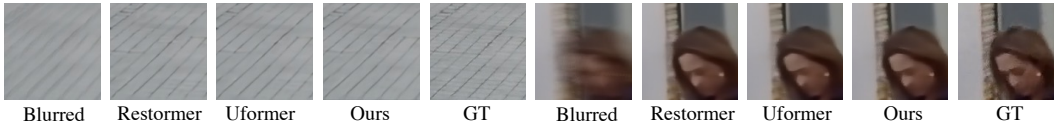}
    \caption{Visual results of global motion deblurring.}
    \label{fig:vis_global}
\end{figure}
\vspace{-1mm}

\section{Further discussion of the components in AdaWPT blocks}
\label{sec:discus}
\subsection{The effectiveness of W-LeFF}
\label{sec:discus_wleff}
To save computational costs as well as inference time, we apply a window-based local-enhanced feed-forward layer (W-LeFF) instead of a locally-enhanced feed-forward layer (LeFF) \citep{wang2022uformer}. In W-LeFF, only the non-pruned windows are processed by a feed-forward mechanism. To ensure a fair comparison between W-LeFF and LeFF, we evaluate them on the all-pruning LMD-ViT network architecture separately. As shown in \autoref{table:leff}, we observe that W-LeFF achieves nearly identical performance to LeFF. Hence, substituting LeFF with W-LeFF does not compromise the local motion deblurring capabilities while simultaneously accelerating the inference speed.
\begin{table}[t]\small
\centering
\caption{The effectiveness of W-LeFF in the proposed LMD-ViT. ``PSNR$_w$'', ``SSIM$_w$'', ``Time'', ``Params'', and ``FLOPs'' denote weighted PSNR, weighted SSIM, inference time, model parameters, and model complexity, respectively.}
\label{table:leff}
\setlength{\tabcolsep}{1mm}
\begin{tabular}{c |c |c | c c c c c c c}
\toprule
No. & Methods & Pruning block & $\uparrow$PSNR & $\uparrow$SSIM & $\uparrow$PSNR$_w$ & $\uparrow$SSIM$_w$ & Time & Params & FLOPs\\
\midrule
1 & LMD-ViT w \textbf{LeFF} & AdaWPT 1\textasciitilde9 &35.45   &0.9288   &30.24 &0.8935 & 0.86s &54.50M &  1.485T \\
2 & LMD-ViT w \textbf{W-LeFF} & AdaWPT 1\textasciitilde9 & 35.42 & 0.9285 & 30.25& 0.8938 &0.56s &54.50M &  1.485T \\
\bottomrule
\end{tabular}
\end{table}

\subsection{The effect of feature channels} \label{sec:derivation}
The number of feature channels also affects the ability of neural networks. With a larger number of feature channels, a neural network can capture more intricate and complex relationships in the input data, resulting in better performance. To verify the capability of our proposed network architecture, we train our network with dimensions 16 and 32 respectively, and compare it with CNN-based LBAG \citep{li2022real} with aligned model parameters, as shown in \autoref{table:dim_resutls}. The comparison between line 3 and line 4 shows an improvement with increased feature channels in LMD-ViT because a larger number of feature channels provides a larger number of meaningful features or attributes, which can be beneficial for window selection and feature manipulation. The comparison between line 2 and line 4 implies that, under approximate model parameters, our proposed model with the adaptive window pruning mechanism is more suitable for the local motion deblurring task with better evaluation scores, fewer model parameters, and faster inference speed. 
\begin{table}[h]\small
\centering
\caption{The effect of feature dimension in the proposed LMD-ViT. ``Feature channels'', ``PSNR$_w$'', ``SSIM$_w$'', ``Time'', ``Params'', and ``FLOPs'' denote the feature channels of each block, weighted PSNR, weighted SSIM, inference time, model parameters, and model complexity respectively.}
\label{table:dim_resutls}
\setlength{\tabcolsep}{0.4mm}
\begin{tabular}{c |c |c |c c c c c c c}
\toprule
No. & Methods & Feature channels & $\uparrow$PSNR & $\uparrow$SSIM & $\uparrow$PSNR$_w$ & $\uparrow$SSIM$_w$  & Time & Params & FLOPs\\
\midrule
1 & LBAG \citep{li2022real}  & 32-64-128 & 34.92 & 0.9318 & 29.30 & 0.8946   & 0.51s & 16.11M & 7.852T\\
2 & LBAG-Large \citep{li2022real} & 60-120-240 & 34.93 & 0.9266 & 29.63 & 0.8871   & 1.13s & 56.58M &17.659T\\
3 & LMD-ViT-Small    & 16-32-64-128-256 & 34.98 & 0.9259 & 29.89 & 0.8907   & 0.23s & 21.59M & 0.311T\\
4 & LMD-ViT       & 32-64-128-256-512 & 35.42 & 0.9285 & 30.25 & 0.8938    & 0.56s & 54.50M & 1.485T\\
\bottomrule
\end{tabular}
\end{table}
\begin{figure}[ht]
    \vspace{-1mm}
    \centering
    \includegraphics[width = \textwidth]{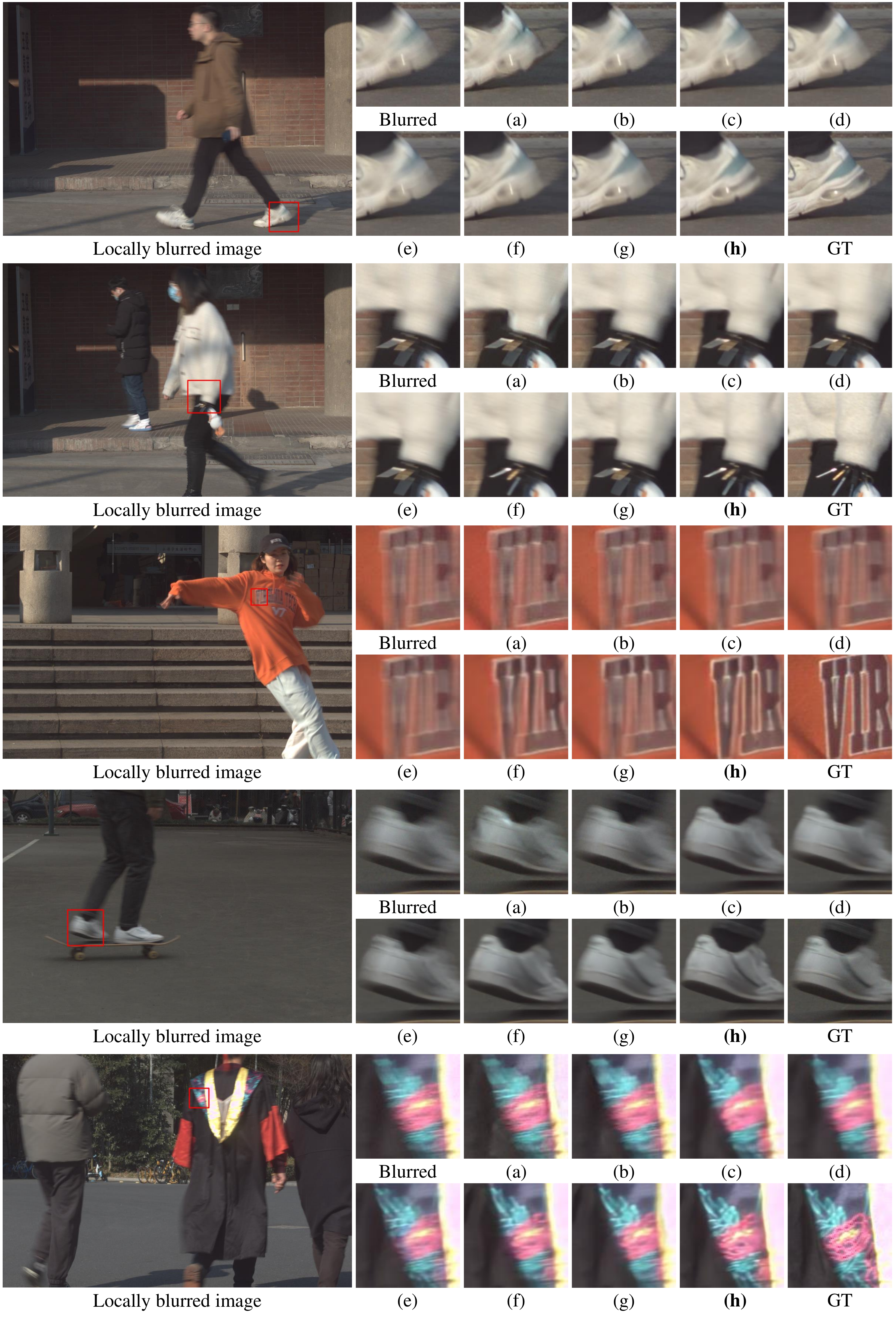}
    \caption{Visual comparing results of state-of-the-art methods for local motion deblurring. (a) DeepDeblur \citep{nah2017deep}; (b) DeblurGAN\_v2 \citep{kupyn2019deblurgan}; (c) HINet \cite{chen2021hinet}; (d) MIMO-UNet \citep{cho2021rethinking}; (e) LBAG \citep{li2022real}; (f) Restormer \citep{zamir2022restormer}; (g) Uformer \citep{wang2022uformer}; (h) LMD-ViT (ours).}
    \label{fig:sub_comp}
\end{figure}
\begin{figure}[ht]
    \vspace{-1mm}
    \centering
    \includegraphics[width = \textwidth]{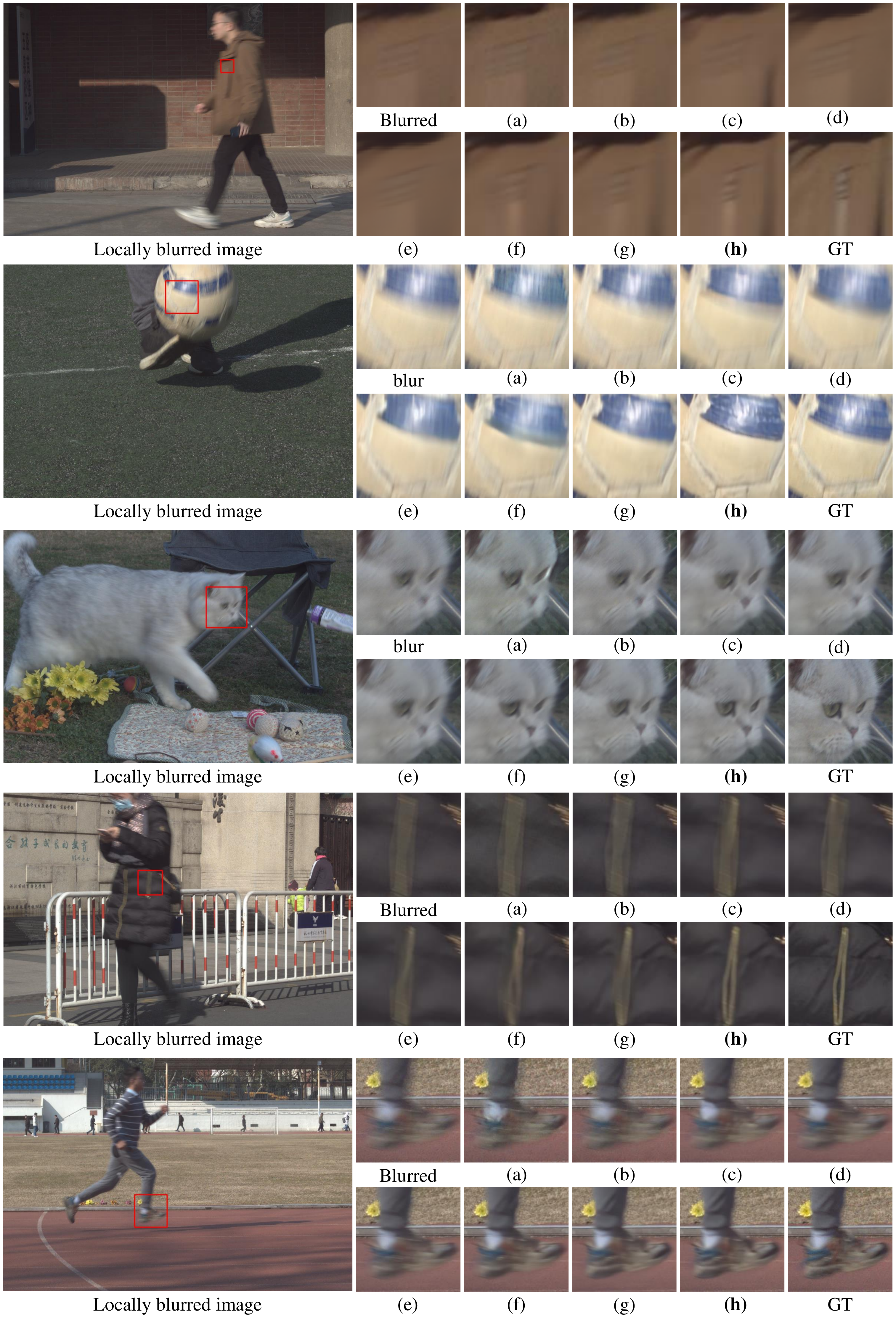}
    \caption{More visual comparing results of state-of-the-art methods for local motion deblurring. The annotations are the same as that in \autoref{fig:sub_comp}.}
    \label{fig:sub_comp2}
\end{figure}
\begin{figure}[ht]
    \vspace{-1mm}
    \centering
    \includegraphics[width = \textwidth]{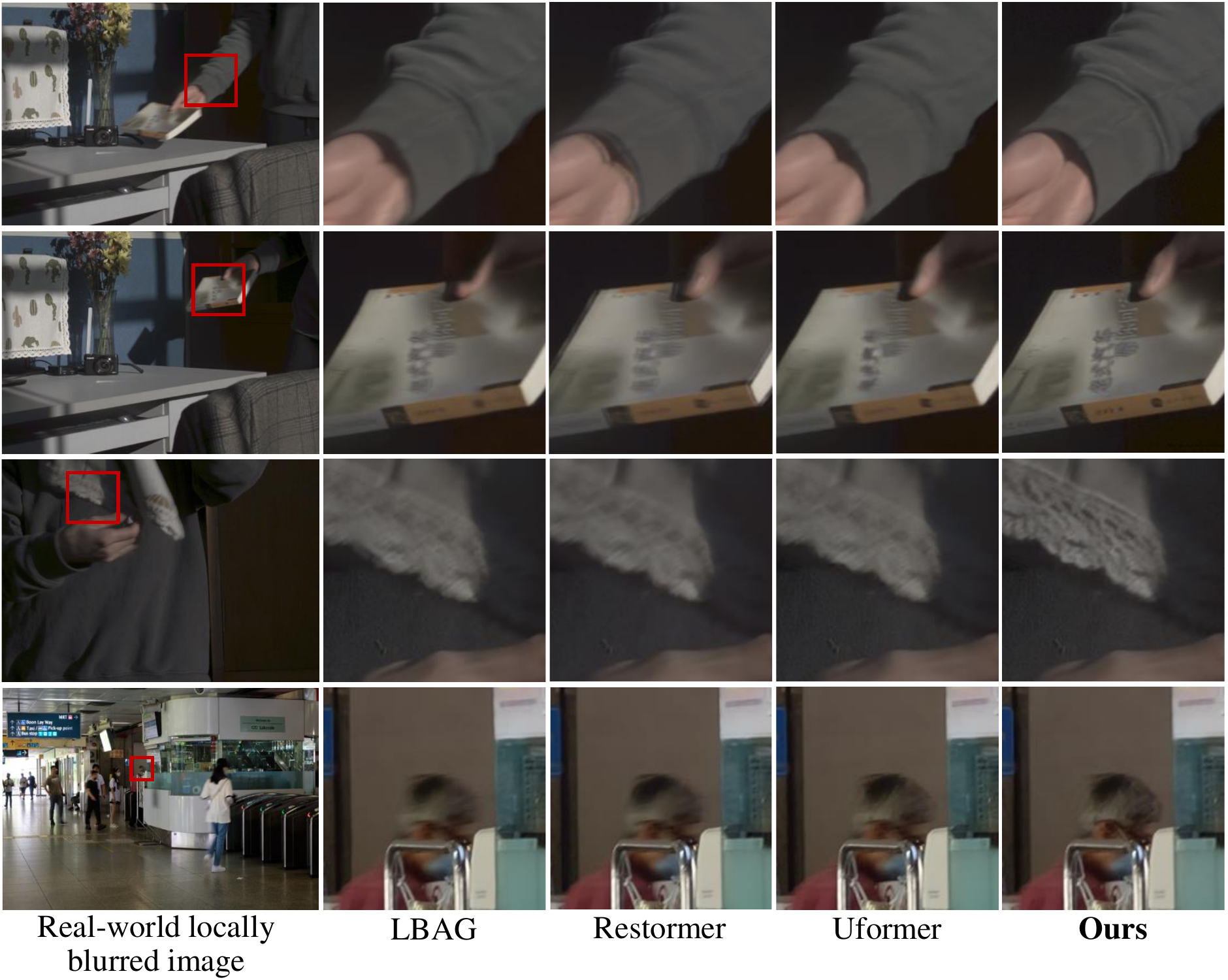}
    \caption{More visual results of user study for real-world local motion deblurring. We compare our proposed LMD-ViT with the top 3 methods, i.e., LBAG~\citep{li2022real},  Restormer \citep{zamir2022restormer}, and Uformer \citep{wang2022uformer}, listed in Table \autoref{table:resutls} of the main paper.}
    \label{fig:sub_usersty}
\end{figure}
\section{More visual results of local motion deblurring}
\label{sec:vis_results}
\subsection{More visual comparisons between LMD-ViT and the compared methods}
We provide more visual results of our proposed LMD-ViT and other baseline methods for local motion deblurring, as shown in \autoref{fig:sub_comp} and \autoref{fig:sub_comp2}. The performance of LMD-ViT surpasses that of other state-of-the-art methods, yielding output images with improved clarity and enhanced details.
\subsection{More visual results of real-world local motion deblurring}
We provide more visual results for real-world local motion deblurring, which are used for conducting user study, as shown in \autoref{fig:sub_usersty}. We conduct a comparison between our method and the top 3 methods listed in Table \autoref{table:resutls} of the main paper. The visual results show that our proposed method exhibits robust deblurring capability and outperforms the state-of-the-art methods on real-world locally blurred images.
\section{Limitations and Broader Impacts}\label{sec:lim_imp}
\noindent\textbf{Limitations} Like other local motion deblurring methods \citep{nah2017deep,kupyn2018deblurgan,kupyn2019deblurgan,chen2021hinet,cho2021rethinking,li2022real,wang2022uformer,zamir2022restormer}, our proposed LMD-ViT is not capable of real-time deblurring. However, we plan to optimize our structure so that it can be applied in real time.

\noindent\textbf{Broader impacts} Our proposed local motion deblurring method democratizes access and empowers users across multiple fields to enhance their visual content, fostering creativity and expression. Firstly, in the realm of consumer electronics and mobile devices, where camera stabilization technology like gimbals has mitigated global motion blur caused by camera shake, local motion blur remains a key problem for static cameras. By incorporating effective local motion deblurring algorithms into these devices, we adaptively enhance image quality with less inference time, leading to superior user experiences and heightened customer satisfaction. In the realm of forensic analysis and surveillance, where local motion blur often plagues static surveillance camera images or video frames, our method enhances the quality of these blurred visuals, enabling better object, individual, and event identification. It proves invaluable in criminal investigations, accident reconstructions, and security-related applications. Last but not least, photographers and visual artists benefit greatly, as it salvages blurred images caused by subject motion, resulting in sharper, visually appealing photographs with enhanced details and clarity. This advancement elevates the overall quality of visual information acquisition and visual arts as expressive mediums.

\end{document}